\documentclass[11pt,a4paper]{article}
\usepackage{acl2015}
\usepackage{times}
\usepackage{url}
\usepackage{latexsym}
\usepackage{textcomp}
\usepackage{graphicx}
\usepackage{algorithm2e}
\usepackage{multirow}
\usepackage{csquotes}
\usepackage{rotating}


\makeatletter
\newcommand{\removelatexerror}{\let\@latex@error\@gobble}
\makeatother

\title{SMPOST: Parts of Speech Tagger for Code-Mixed Indic Social Media Text}

\author{Deepak Gupta \\
  Dept. of CSE\\
  IIT Patna, India \\
  {deepak.pcs16@iitp.ac.in} \\\And
  Shubham Tripathi \\
  Dept. of EE \\
  MNIT Jaipur, India \\
  {stripathi1770@gmail.com} \\\And
   Asif Ekbal, Pushpak Bhattacharyya \\
  Dept. of CSE\\
  IIT Patna, India \\
  {\{asif, pb\}@iitp.ac.in} \\}

\date{}

\begin{document}
\maketitle
\begin{abstract}
Use of social media has grown dramatically fast during the past few years. Users usually follow informal languages in communicating through social media. This language of communication is often mixed in nature, where people transcribe their regional language with English. This technique of writing is increasing its popularity rapidly. 
Natural language processing (NLP) aims to infer the information from these text where Part-of-Speech (PoS) tagging plays an important role in getting the prosody of the written text. For the task of PoS tagging on \textit{Code-Mixed Indian Social Media Text}, we develop a supervised system based on Conditional Random Field classifier. In order to tackle the problem effectively, we have focused on extracting rich linguistic features. 
We participate in three different language pairs, ie. English-Hindi, English-Bengali and English-Telugu on three different social media platforms, Twitter, Facebook \& WhatsApp. The proposed system is able to successfully assign coarse as well as fine grained PoS tag labels for a given a code-mixed sentence. 
Experiments show that our system is quite generic and shows encouraging performance levels on all the three language pairs in all the domains.
\end{abstract}
\section{Introduction}
Code-mixing refers to the mixing of two or more languages or language varieties. The terms, code switching and code mixing, are nowadays interchangeably used. The sheer access to internet has resulted in increased social media involvement among the masses. Over the past decade, Indian language contents on various media types such as blogs, email, websites and chats have increased significantly. With the advent of smart phones, more people have now access to social media through WhatsApp, Twitter, Facebook via which they share their opinions on people, products, services, organizations and governments. This abundance of social media data has created many new opportunities for information access, but has also led to many novel challenges.

Part-of-Speech (PoS) tagging is a fundamental task of Natural Language Processing.  Many higher level NLP tasks require input as a PoS-tagged sentence for parsing. The result of PoS tagging on English-Spanish code-mixed data have been reported in \cite{solorio2008part}. Recently, \cite{vyas2014pos} has proposed an architecture of PoS tagging for English-Hindi code-mixed data. A language identification system developed by \cite{gella2013query} used a simple heuristic approach to form chunks of data for the same language. They apply a CRF-based technique for Hindi and Twitter POS tagger \cite{owoputi2013improved} for English, and map it to the Universal tagset \cite{petrov2011universal}.
This paper explores the task of PoS tagging in code-mixed Indian social media text using a supervised learning approach. The task is to perform PoS tagging in three language pair data sets, namely English-Hindi (EN-HI), English-Bengali (EN-BE) and English-Telugu (EN-TE). PoS tagging for all the three types of data, i.e. Facebook, Twitter and WhatsApp, have been divided in Rule and Classifier based tagging.
At first we build a rule-based PoS tagger to identify a subset of PoS tags, followed by a CRF based classification system for the others. In our literature survey on PoS tagging on social code-mixed data, we observe that Universal PoS tags \cite{petrov2011universal} are the most popular. On the other hand, the data sets provided for this shared task were PoS tagged using BIS tagset\cite{jamatia2015part}. For the \textit{constrained system}, we used the respective language pair data set provided by the organizer \cite{POSicon2016}. We used the ICON-2015 PoS tagging shared task data in addition to this year data to build the \textit{unconstrained system}.
The source code of our system can be found here.\footnote{\url{https://github.com/stripathi08/pos_cmism}}
\section{Rule based PoS Tagging}
Social media data is an amalgamation of diverse heuristics. The task of developing systems for the prediction of PoS is, thus immensely complex. Prior to sending word tokens for classification, it is necessary to remove the inappropriate tokens. This includes all residuals, numerals, emoticons, website links and twitter-specific word tokens like `@',`\#',`\texttildelow' and many more. In order to address the variations of such tokens in the testing data, we develop a rule-based system for sequentially removing one type after the another.\\
We started by removing all the punctuations and their respective extensions such as `\textbf{....}', `\textbf{!.!.}' and `\textbf{\texttildelow}'. All the numeric tokens were tackled next. This includes tokens starting with a numeric entry and ending with `\textit{st}', `\textit{nd}', `\textit{rd}', `\textit{th}' for tagging tokens like `\textit{1st}', `\textit{2nd}', `\textit{3rd}' and `\textit{100th}', respectively. Cellular numbers and textual representations such as `\textit{lakh}' and `\textit{million}' were also tagged as numerals. Web URLs were searched through a regular expression based classification system, capturing tokens having strings such as `\textit{.com}', `\textit{.me}', `\textit{.org}', `\textit{.in}' and starting with `\textit{https://}', `\textit{http://}'. \\
Emoticons were searched via a two-step mechanism: a regular expression based system followed by a dictionary look-up. The tokens were first checked for common emoticons such as `\textit{:)}', `\textit{:(}' and then sent for a match from a dictionary of popular emoticons. Any token with a unicode expression was marked as `\textit{RD\_UNK}'. Rest of the tokens are considered to form a set of final cleaned data. We followed the same processes for both training and test data sets. Features are extracted on these datasets for classifier's training. 
The statistics of the tag-set which was handled through rule based approach, are given in  Table-\ref{rule-stat}. The details steps of rule based tagging are described in Algorithm-\ref{post-rule-en-hi}.
\begin{table}[]
\centering
\resizebox{\linewidth}{!}{%
\begin{tabular}{|c|c|c|c|c|}
\hline
\begin{turn}{45}\textbf{Category}\end{turn} & \textbf{POS Tag} & \textbf{English-Hindi} & \textbf{English-Bengali} & \textbf{English-Telugu} \\ \hline
\multirow{6}{*}{\textbf{\begin{turn}{90}Coarser\end{turn}}} & \textbf{G\_X} & 5844 & 2052 & 6616 \\ \cline{2-5} 
 & \textbf{E} & 344 & 104 & 320 \\ \cline{2-5} 
 & \textbf{@} & 993 & 269 & 1278 \\ \cline{2-5} 
 & \textbf{U} & 300 & 7 & 322 \\ \cline{2-5} 
 & \textbf{\$} & 439 & 100 & 323 \\ \cline{2-5} 
 & \textbf{\#} & 587 & 118 & 278 \\ \hline
\multirow{8}{*}{\textbf{\begin{turn}{90} Finer\end{turn}}} & \textbf{RD\_PUNC} & 4544 & 1781 & 3205 \\ \cline{2-5} 
 & \textbf{E} & 344 & 104 & 320 \\ \cline{2-5} 
 & \textbf{\#} & 587 & 118 & 278 \\ \cline{2-5} 
 & \textbf{U} & 300 & 7 & 322 \\ \cline{2-5} 
 & \textbf{@} & 993 & 269 & 1278 \\ \cline{2-5} 
 & \textbf{\$} & 439 & 100 & 323 \\ \cline{2-5} 
 & \textbf{RD\_SYM} & 266 & 48 & 2 \\ \cline{2-5} 
 & \textbf{RD\_UNK} & 15 & 0 & 50 \\ \hline
\end{tabular}%
}
\caption{Statistics of POS tags from Training data, handled through Rule based tagging}
\label{rule-stat}
\end{table}
\begin{algorithm}
\SetKwInOut{Input}{Input}\SetKwInOut{Output}{Output}
  \Input{A word $w$ from sentence $S$}
\Output{ Finer level and Coarser level POS tag: F-TAG, C-TAG }
  \uIf{isPunctuation(w)}{
  F-TAG=$\mathbf{RD\_PUNC}$\;
  C-TAG=$\mathbf{G\_X}$\;
   }
  \uElseIf{isContains(w,`\texttildelow')}{
          F-TAG=$\mathbf{RD\_SYM}$\;
  C-TAG=$\mathbf{G\_X}$\;
        }
        \uElseIf{isContainsUnicode(w)}{
             F-TAG=$\mathbf{RD\_UNK}$\;
  C-TAG=$\mathbf{G\_X}$\;
        }
         \uElseIf{isDigit(w)}{
             F-TAG=$\mathbf{\$}$\;
  C-TAG=$\mathbf{\$}$\;
        }
          \uElseIf{isDigit(w) AND ( isEndsWith(w,\enquote{st}) OR isEndsWith(w,\enquote{nd}) OR isEndsWith(w,\enquote{rd}) )}{
             F-TAG=$\mathbf{\$}$\;
  C-TAG=$\mathbf{\$}$\;
        }
        \uElseIf{isStartsWith(w,`+91') OR isInNumberDictionary(w)}{
             F-TAG=$\mathbf{\$}$\;
  C-TAG=$\mathbf{\$}$\;
        }
         \uElseIf{ isEndsWith(w,\enquote{.com}) OR isEndsWith(w,\enquote{.org}) OR isEndsWith(w,\enquote{.me}) }{
             F-TAG=$\mathbf{U}$\;
  C-TAG=$\mathbf{U}$\;
        }
   \uElseIf{ isStartsWith(w,\enquote{http://}) OR isStartsWith(w,\enquote{https://}) OR isStartsWith(w,\enquote{www://}) }{
             F-TAG=$\mathbf{U}$\;
  C-TAG=$\mathbf{U}$\;
        }
         \uElseIf{isInEmoticonsDictionary(w)}{
             F-TAG=$\mathbf{E}$\;
  C-TAG=$\mathbf{E}$\;
        }
        \uElseIf{isStartsWith(w,`@')}{
             F-TAG=$\mathbf{@}$\;
  C-TAG=$\mathbf{@}$\;
        }
         \uElseIf{isStartsWith(w,`\#')}{
             F-TAG=$\mathbf{\#}$\;
  C-TAG=$\mathbf{\#}$\;
        }
         
        \uElse{
            F-TAG=$\mathbf{?}$\;
  C-TAG=$\mathbf{?}$\;
        }             
\textbf{return} F-TAG, C-TAG;
\label{post-rule-en-hi}
\caption{Algorithm for Rule based tagging, if algorithm return `?' as output then the word $w$ is send to CRF classifier in sequence with respect to the sentence $S$.}
\end{algorithm}      
\section{Feature Extraction}\label{feature}
The proposed system uses an exhaustive set of features for PoS labelling. The features are explained in brief below:
\begin{enumerate}
\item \textbf{Context word}: Local contextual information is useful to determine the type of the current word. We use the contexts of previous two and next two words as features. 
\item \textbf{Character n-gram}: 
Character n-gram is a contiguous sequence of n characters extracted from a given word. The set of n-grams that can be generated for a given token is basically the result of moving a window of n characters along the text. We extracted character n-grams of length \textit{one} (unigram), \textit{two}(bigram) and \textit{three} (trigram), and use these as features of the classifiers.
\item \textbf{Word normalization }: Words are normalized in order to capture the similarity between two different words that share some common properties. Each uppercase letter is replaced by \lq A\rq, lowercase by 'a' and number by '0'.
\begin{table}[h]
\centering
\begin{tabular}{|c|c|}
\hline
\textbf{Words} & \textbf{Normalization} \\ \hline
NH10  & AA00          \\ \hline
Maine & Aaaaa         \\ \hline
NCR   & AAA           \\ \hline
\end{tabular}
\end{table}
\item \textbf{Prefix and suffix}: Prefix and suffix of fixed length character sequences (here, 3) are stripped from each token and used as features of the classifier. 

\item \textbf{Word class feature}: This feature was defined to ensure that the words having similar structures belong to the same class. In the first step we normalize all the words following the process as mentioned above. Thereafter, consecutive same characters are squeezed into a single character. For example, the normalized word \textit{AAAaaa} is converted to \textit{Aa}. We found this feature to be effective for the biomedical domain, and we directly adapted this without any modification.
\item \textbf{Word position:} In order to capture the word context in the sentence, we have used a numeric value to indicate the position of word in the sentence. The normalized position of word in the sentence is used as a features. The feature values lies in the ranges between 0 and 1. 
\item \textbf{Number of upper case characters:} This feature takes into account the number of uppercase alphabets in the word. The feature is relative in nature and ranges between $0$ and $1$. 
\item \textbf{Word probability:} This feature finds the probability of a word to be labeled with the same as in training data. The length of this feature vector is the total number of labels or output tags, where each bit represents an output tag, initialized with $0$. If the word does not appear in training, each bits retain their initially marked value $0$. Based on the probability value, we have two features:
\begin{enumerate}
\item \textbf{Top@1-Probability:} The probability was calculated for current word $w$ labeled as POS tag $t$ in training set. For the output tag with highest probability, its corresponding bit in the feature vector is set to $1$. All other bits remain as $0$.
\item \textbf{Top@2-Probability:}  The probability was calculated for current word $w$ labeled as POS tag $t$ in training set. For the output tag with highest and second highest probability, their corresponding bit are set to $1$ in the feature vector. All other bits remain as $0$.
\end{enumerate}
\item \textbf{Binary features:} 
We define the following binary-valued features from the information available in the training data.
\begin{enumerate}
\item \textbf{isSufficientLength:} Since most of the entity from training data have a significant length. Therefore we set a binary feature to fire when the length of token is greater than a specific threshold value. The threshold value $4$ is used to extract the binary features. 
\item \textbf{isAllCapital:} This value of this feature is set when all the character of current token is in uppercase.
\item \textbf{isFirstCharacterUpper:} This value of this feature is set when the first character of current token is in uppercase..
\item \textbf{isInitCap}: This feature checks whether the current token starts with a capital letter or not. This provides an evidence for the target word to be of NE type for the English language. 
\item \textbf{isInitPunDigit}: We define a binary-valued feature that checks whether the current token starts with a punctuation or a digit. It indicates that the respective word does not belong to any language. Few such examples are \textit{4u,:D}, \textit{:P} etc. 
\item \textbf{isDigit}: This feature is fired when the current token is numeric. 
\item \textbf{isDigitAlpha}: We define this feature in such a way that checks whether the current token is alphanumeric. The word for which this feature has a true value has a tendency of not being labeled as any named entity type. 
\item \textbf{isHashTag}: Since we are dealing with tweeter data , therefore we encounter a lot of hashtag is tweets. We define the binary feature that checks whether the current token starts with \# or not.
\end{enumerate}
\item \textbf{Stemming:} For \textit{English} Language tokens, this features separately adds the stemmed  version of the token as a feature. We have used the Porter Stemmer algorithm, from the NLTK tool\footnote{http://www.nltk.org/}, to get the stemmed version. For \textit{Non-English} tokens, this feature is set to null.
\item \textbf{Phonetic normalization:}  For \textit{English} Language tokens, this features adds the Phonetics of the token as a feature. We have used the Double Metaphone phonetic matching algorithm \cite{philips2000double} to extract the feature. For \textit{Non-English} tokens, this feature is set to null.
\end{enumerate}

\begin{table*}[t]
\centering
\resizebox{\linewidth}{!}{%
\begin{tabular}{ |c|c|c|c|c|c|c|c|c|c|c| }
 \hline
 \multicolumn{11}{|c|}{Fine-Grained} \\
 \hline
 Runs&Measures&BN-FB&BN-TWT&BN-WA & HI-FB&HI-TWT&HI-WA & TE-FB&TE-TWT&TE-WA \\
 \hline
 \multirow{3}{1pt}{1}&P&0.664&0.632&0.739 &  0.583&0.782&0.67 &  0.68&0.696&0.742 \\
 					&R&0.999&0.632&0.997 &  0.945&0.987&0.987 &  0.997&0.995&0.978 \\
 					&F&0.797&0.632&0.848 &  0.721&0.872&0.798 &  0.807&0.819&0.843 \\
 \hline
 \multirow{3}{1pt}{2}&P&0.712&0.65&0.727 &  0.678&0.777&0.602 &  0.656&0.638&0.651 \\
 					&R&0.999&0.65&0.727 &  0.997&0.986&0.992 &  0.997&0.991&0.998 \\
 					&F&0.831&0.605&0.727 &  0.807&0.869&0.748 &  0.791&0.776&0.788 \\
 \hline
 \end{tabular}
 }
\caption{Precision (P), Recall (R) and F-Scores (F) of constrained system, Runs-1 and Run-2 on Finer grained tag set. Here the notation are, \textbf{BN-FB}: Facebook data in English-Bengali, \textbf{BN-TWT}: Twitter data in English-Bengali, \textbf{BN-WA}: WhatsApp data in English-Bengali, \textbf{HI-FB}: Facebook data in English-Hindi, \textbf{HI-TWT}: Twitter data in English-Hindi, \textbf{HI-WA}: WhatsApp data in English-Hindi, \textbf{TE-FB}: Facebook data in English-Telugu, \textbf{TE-TWT}: Twitter data in English-Telugu, \textbf{TE-WA}: WhatsApp data in English-Telugu }
\label{table-1}
\end{table*} 

\begin{table*}[t]
\centering
\resizebox{\linewidth}{!}{%
\begin{tabular}{ |c|c|c|c|c|c|c|c|c|c|c| }
 \hline
 \multicolumn{11}{|c|}{Coarse-Grained} \\
 \hline
 Runs&Measures&BN-FB&BN-TWT&BN-WA & HI-FB&HI-TWT&HI-WA & TE-FB&TE-TWT&TE-WA \\
 \hline
 \multirow{3}{1pt}{1}&P&0.75&0.702&0.775 &  0.627&0.832&0.773 &  0.734&0.738&0.787 \\
 					&R&0.75&0.702&0.775 &  0.627&0.989&0.773 &  0.999&0.997&0.995 \\
 					&F&0.750&0.702&0.775 &  0.627&0.904&0.773 &  0.846&0.848&0.879 \\
 \hline
 \multirow{3}{1pt}{2}&P&0.788&0.722&0.769 &  0.751&0.824&0.692 &  0.707&0.23&0.615 \\
 					&R&0.788&0.722&0.769 &  0.751&0.989&0.692 &  0.997&0.23&0.615 \\
 					&F&0.788&0.722&0.769 &  0.751&0.899&0.692 &  0.827&0.230&0.615 \\
 \hline

 \end{tabular}
 }
\caption{Precision (P), Recall (R) and F-Scores (F) of constrained system, Runs-1 and Run-2 on Coarser grained tag set, Notations are the same as described in Table-2}
\label{table-2}
\end{table*} 
\section{Data Set \& Experimental Setup}
\subsection{Dataset}
The data sets used in the experiment are provided by the organizer of POS tagging tool contest at ICON-2016 \cite{POSicon2016}. Data sets consist of Facebook, Twitter and WhatsApp utterances at fine grained and coarse grained levels. There are three different code-mixed language pair English-Hindi, English-Bengali and English-Telugu. Data sets from each three major social media platforms, Facebook, Twitter and WhatsApp, are provided to build the system. More details about the data sets can be found in the overview paper \cite{POSicon2016}. 
\subsection{Experiment}
The feature set discussed in Section-\ref{feature} were used to build a POS model. Conditional random field (CRF) is used as the underlying classifier. $CRF^{++}$\footnote{https://taku910.github.io/crfpp/}, an implementation of CRF is used to perform the experiment. We have used the default setting of $CRF^{++}$ throughout the experiment. As $CRF^{++}$ uses a specified feature template, therefore to find the optimal feature template a series of experiments were performed on the training data set in a cross-validated manner. However, we tune the feature template on English-Hindi data set only and use the optimal template for all the three language pairs. 
\section{Results}
We have submitted two separate constrained runs for this shared task. The runs differ in the manner of their training while the classification system remains the same. The description of  our constrained runs are as follows:
\begin{enumerate}
\item \textbf{Run-1}: The system was trained by augmenting the respective language pair  training data from all three social media platform \textit{viz} Facebook, Twitter and WhatsApp. Therefore the same model was used to get the POS tag of Facebook, Twitter and WhatsApp test data. The same strategy was followed for both coarse grained and fine grained POS tag. 
\item \textbf{Run-2}:  The system was trained individually on only respective language pair training data from different social media platforms \textit{viz} Facebook, Twitter and WhatsApp. Therefore the model, trained on Facebook, twitter and WhatsApp training data was used to get the POS tag of Facebook, Twitter and WhatsApp test data respectively. The same strategy was followed for both coarser grained and finer grained POS tag. 
\end{enumerate}
The detailed results of both the runs on finer and coarse grained POS tagging are shown in Table-\ref{table-1} and Table-\ref{table-2}.\\
The unconstrained system is an extension of Run-1 of constrained system along with previous year's data. We have combined the respective language pair training data from previous year's task and this year's task.  While cross-validating to find the optimal system performance, we could not observe a significant increase in the performance.
It is observed from the results of both the constrained runs that, in general, Run-1 performed better in WhatsApp domain among all the language pairs in the finer as well as coarser grained tag set. Similarly, Run-1 performed better in Twitter domain too, except for the Bengali-English language pair. Conversely, Run-2 performed better in Facebook domain among all the language pairs in both the tag sets.    
\section{Future work and Conclusion}
This paper describes the approach for Parts of Speech tagging at finer and coarser level on Bengali-English, Hindi-English and Telugu-English language pairs code-mixed data. The system performed exceptionally well in all the domains for finer and coarser tag set, barring the Telugu Twitter Constrained Run-2 on Coarse grained tag set. We are working on the application of Deep Neural Networks in the task of POS tagging on code-mix environment. A lack of labeled data has been the major cause for poor results when applying Recurrent Neural Networks and Long-Short Term Memory (LSTM) models for the classification task. We are currently working on combining a CRF based classifier system and a Bi-Directional LSTM (BLSTM) model for improving the results on code-mixed data sets.
\bibliography{bibliography}
\bibliographystyle{acl}
\end{document}